\documentclass[11pt,a4paper]{article}

\usepackage[utf8]{inputenc}
\usepackage[english]{babel}
\usepackage[margin=1in]{geometry}
\usepackage{graphicx}
\usepackage{natbib}
\usepackage{url}
\usepackage{amsmath}
\usepackage{amssymb}
\usepackage{caption}
\usepackage{algorithm}
\usepackage{algorithmic}
\usepackage{newfloat}
\usepackage{listings}
\usepackage{hyperref}

\newcommand{\ar}[1]{\textit{#1}}

\DeclareCaptionStyle{ruled}{labelfont=normalfont,labelsep=colon,strut=off}
\floatstyle{ruled}
\newfloat{listing}{tb}{lst}{}
\floatname{listing}{Listing}

\hypersetup{
    pdfauthor={Saad Mankarious and Ayah Zirikly},
    pdftitle={},
    pdfkeywords={Arabic NLP, mental health, dataset, automatic annotation, social media}
}

\title{\textbf{\Large Style Transfer as Bias Mitigation: Diffusion Models for Synthetic Mental Health Text for Arabic}}

\author{
    Saad Mankarious \quad Ayah Zirikly \\
    School of Engineering and Applied Science \\
    George Washington University, Washington, D.C. 20037 \\
    \texttt{\{saadm, ayah.zirikly\}@gwu.edu}
}

\date{\today}

\begin{document}

\noindent\makebox[\linewidth]{\rule{\textwidth}{2pt}}\\[0.5em]
\begin{center}
{\Large\textbf{Style Transfer as Bias Mitigation: Diffusion Models for Synthetic Mental Health Text for Arabic}}\\[0.5em]
Saad Mankarious \quad Ayah Zirikly \\
School of Engineering and Applied Science \\
George Washington University, Washington, D.C. 20037 \\
\texttt{\{saadm, ayah.zirikly\}@gwu.edu}
\end{center}
\noindent\makebox[\linewidth]{\rule{\textwidth}{2pt}}\\[1.5em]

\begin{abstract}
Synthetic data offers a promising solution for mitigating data scarcity and demographic bias in mental health analysis, yet existing approaches largely rely on pretrained large language models (LLMs), which may suffer from limited output diversity and propagate biases inherited from their training data. In this work, we propose a pretraining-free diffusion-based approach for synthetic text generation that frames bias mitigation as a style transfer problem. Using the CARMA Arabic mental health corpus, which exhibits a substantial gender imbalance, we focus on male-to-female style transfer to augment underrepresented female-authored content. We construct five datasets capturing varying linguistic and semantic aspects of gender expression in Arabic and train separate diffusion models for each setting. Quantitative evaluations demonstrate consistently high semantic fidelity between source and generated text, alongside meaningful surface-level stylistic divergence, while qualitative analysis confirms linguistically plausible gender transformations. Our results show that diffusion-based style transfer can generate high-entropy, semantically faithful synthetic data without reliance on pretrained LLMs, providing an effective and flexible framework for mitigating gender bias in sensitive, low-resource mental health domains.
\end{abstract}

\section{Introduction}
The use of text-based data in mental health analysis is important for the diagnosis and study of several mental health conditions \cite{zirikly2019clpsych, cohan2018smhd, mankarious2025carma}. However, the availability of such textual content is constrained due to the difficulty of obtaining sensitive data under privacy considerations, as well as the biased nature of the media from which most of this text is collected. Social media platforms, for example, often exhibit disproportionate representation across demographic groups, such as an overrepresentation of males relative to females and biases toward specific populations \cite{lozoya2023identifying, torralba2011unbiased, varshney2019pretrained}.

Generative models present a promising approach for producing high-quality synthetic text data that can be useful for a wide range of tasks \cite{stevenson2022putting}. Synthetic text data offers an opportunity to address gender bias in mental health analysis by enabling the augmentation of underrepresented classes, thereby supporting fairer and more balanced downstream models \cite{lozoya2023identifying}. However, text-based synthetic data generation has largely been limited to the use of large language models (LLMs) trained on massive internet-scale corpora. This reliance introduces two critical limitations. First, sampling a high-entropy corpus from an LLM is challenging due to a phenomenon known as model collapse, where the model fails to adequately regulate output diversity as a result of its extensive pretraining. Second, LLM outputs often reflect inherited biases present in their pretraining data, as demonstrated in prior work \cite{lozoya2023identifying, lambrecht2019algorithmic, buolamwini2018gender}.

We propose an alternative approach for generating synthetic data to address gender bias in mental health analysis using a pretraining-free diffusion-based architecture. We formulate the synthetic data generation task as a style transfer problem, in which generation is conditioned on samples drawn from real data. This allows us to generate counterparts of the limited available data while maintaining high data entropy. For specificity, we focus on addressing gender bias in an Arabic mental health dataset. Given a dataset skewed toward male-authored posts, our goal is to augment the female class by generating equivalents of male posts written in a female style. A key aspect of our approach is preserving the original data distribution by conditioning generation entirely on content from the same dataset. This design yields two important advantages over LLM-based approaches. First, it supports high-entropy generation, as each synthetic sample is conditioned on an existing data point, assuming the original dataset itself exhibits sufficient diversity. Second, it avoids reliance on pretrained weights, thereby reducing the risk of propagating external biases introduced through large-scale internet pretraining.

To the best of our knowledge, no prior work has addressed synthetic text data generation without relying on a pretrained LLM. We summarize our contributions as follows:
\begin{itemize}
\item We produce five datasets with varying levels of semantic and linguistic complexity to support male-to-female style transfer in Arabic text.
\item We train five pretraining-free diffusion models capable of mimicking female writing style and generating female-styled content conditioned on male-authored posts.
\item We demonstrate the utility of the generated synthetic data by showing high entropy relative to the original corpus and clear stylistic shifts from male to female writing.
\end{itemize}

\section{Previous Work}

\subsection{Synthetic Data in the Medical Field}
Data augmentation in the health domain has traditionally been dominated by techniques such as Synthetic Minority Oversampling (SMOTE) and Synthetic Minority Augmentation (SMA) \cite{juwara2024evaluation}. While these approaches are effective in mitigating quantitative data scarcity, they fail to capture the complexity of text-based data, as they operate in the feature space rather than directly in the data space. This reliance on feature-level augmentation limits their ability to address more nuanced sources of bias, such as gender bias, which manifest through linguistic and contextual patterns.

To capture more sophisticated forms of bias, including gender bias in clinical settings, researchers have increasingly turned to large language models (LLMs). These models leverage extensive pretraining corpora and strong reasoning capabilities to generate fluent natural language. However, the pretraining data used by LLMs often contains embedded biases that may be reflected or amplified in generated outputs. For example, \cite{lozoya2023identifying} provide evidence of gender bias when prompting GPT-3 to generate text in male versus female voices. Furthermore, prior studies have shown that such biases can lead to harmful outcomes in mental health analysis \cite{buolamwini2018gender, lambrecht2019algorithmic}. Additional work demonstrates that LLMs frequently portray women as less powerful than men and associate Muslim groups with violence \cite{abid2021persistent, lucy2021gender}.

\subsection{Diffusion Models for Text Generation}
Diffusion models, originally introduced by \cite{sohl2015deep} in 2015, have proven effective for recovering latent data distributions by progressively adding noise and then learning to reverse this process, particularly in image generation tasks. More recently, diffusion-based approaches have been extended to continuous text generation, most notably by \cite{gong2022diffuseq}. DiffuSeq is a pretraining-free diffusion-based architecture designed for text-to-text generation and style transfer. \cite{gong2022diffuseq} report that DiffuSeq achieves text generation performance comparable to fine-tuned GPT-2 base and large models, despite being non-autoregressive. In this work, we adapt the model proposed by \cite{lyu2023fine} to generate synthetic data that augments scarce female-authored posts by transforming male-authored posts into a female writing style.

\section{Data}

\subsection{CARMA Mental Health Corpus}
We work with the CARMA dataset \cite{mankarious2025carma}, an Arabic corpus curated from social media that encompasses six mental health conditions in addition to a control group. The corpus was constructed using self-reported diagnoses \cite{cohan2018smhd}, in which users disclose their mental health conditions; these disclosures are treated as ground truth in the absence of clinical confirmation. The dataset exhibits a significant gender imbalance, with a female-to-male ratio of approximately one to three. Table~\ref{tab:desc-stats} presents additional statistics for the selected portions of the corpus.

\begin{table}
\centering
\caption{User and post statistics of the two groups selected from the CARMA corpus.}
\label{tab:desc-stats}
\begin{tabular}{lccccc}
\hline
\textbf{Condition} & \textbf{\# Users} & \textbf{\# Posts} & \textbf{Avg. Post Length} & \textbf{\# Females} & \textbf{\# Males} \\
\hline
Control & 4,086 & 270,622 & 44.87 & 625 & 2,026 \\
Depression & 171 & 21,085 & 49.32 & 44 & 127 \\
\hline
\end{tabular}
\end{table}

\subsection{Five Datasets for Male-to-Female Style Transfer}
Our objective is to transfer linguistic style to better represent the minority female class using the male-authored portion of the dataset. To this end, we construct five datasets with increasing levels of sophistication, each designed to capture different aspects of female voice expression in Arabic. These datasets aim to model a range of linguistic and semantic phenomena through which gender may be expressed in Arabic, ranging from simple pronoun switching (given the gendered nature of the language) to more complex transformations involving affect and emotional intensity. The datasets were developed in consultation with two native Arabic speakers with expertise in language use and expression. Table~\ref{tab:datasets} summarizes the five datasets and provides illustrative examples. Each dataset is used to train a separate diffusion model.

We leverage the \texttt{gpt-4o} model to generate the datasets. The Appendix details the prompts used for generating each dataset. We use a temperature of 0 for reproducibility and employ the batching API for more efficient annotation.

Table~\ref{tab:dataset-stats} reports statistics for the five generated datasets. We aim to match a target size of XX instances per dataset. Table~\ref{tab:datasets} further presents examples illustrating the semantic and syntactic role of each dataset in the context of gender style transfer.

\subsubsection{D1: Pronoun Switching}
This dataset represents a light, straightforward transformation in which only pronoun forms are strictly switched from male to female\footnote{In Arabic, personal pronouns differ by gender.}.  

\subsubsection{D2: Adjectival Lexical Preference Shift}
This dataset emphasizes the use of adjectives and stronger adjectival forms.

\subsubsection{D3: First-Person Self-Expression Framing}
This category focuses on increased use of personal language and self-referential expressions.

\subsubsection{D4: Politeness and Hedging Pragmatic Shift}
In this dataset, sentences originally written in a male voice are rewritten to reflect a more polite style, with greater use of hedging and pragmatic softening.

\subsubsection{D5: Emotion and Affect Density Amplification}
This dataset aims to intensify emotional expression more broadly, rather than strengthening specific adjectives as in D2.

\subsection{Statement of Ethics}
We acknowledge that the design of the five datasets may appear problematic at first glance, as it could be interpreted as reinforcing stereotypical associations of female language with attributes such as modesty, politeness, or emotionality. Based on our expertise in Arabic linguistic expression, we recognize that such categorizations are not universally valid, despite observing evidence of distinct linguistic patterns that align with these categories, as discussed in the Appendix. However, to simplify the style transfer task, we adopt these categorizations as an initial framework. Throughout this work, we maintain a clear separation between the design of the style transfer task and the underlying data. Consequently, the methodology presented in the subsequent section is agnostic to the specific source and target styles and can be applied to alternative configurations in future applications.

\begin{table}
\centering
\begin{tabular}{lccccc}
\hline
\textbf{Dataset} & \textbf{\#Posts} & \textbf{Src Avg Len} & \textbf{Src Std} & \textbf{Trg Avg Len} & \textbf{Trg Std} \\
\hline
D1 & 2265 & 34.15 & 33.03 & 25.89 & 24.81 \\
D2 & 2265 & 34.15 & 33.03 & 26.76 & 23.84 \\
D3 & 2265 & 34.15 & 33.03 & 26.55 & 23.84 \\
D4 & 2265 & 34.15 & 33.03 & 32.27 & 24.39 \\
D5 & 2265 & 34.15 & 33.03 & 30.98 & 22.72 \\
\hline
\end{tabular}
\caption{Dataset statistics aggregated across all splits. Source (Src) and target (Trg) lengths are measured in number of words.}
\label{tab:dataset-stats}
\end{table}

\begin{table}
\centering
\renewcommand{\arraystretch}{1.25}
\begin{tabular}{p{0.20\textwidth} p{0.34\textwidth} p{0.22\textwidth} p{0.22\textwidth}}
\hline
\textbf{Dataset} & \textbf{Description} & \textbf{Source} & \textbf{Target} \\
\hline

D1: Strict Morphosyntactic Gender Agreement &
Only grammatical gender markers are changed; meaning and sentiment remain identical. &
\ar{Ana kunt muttawattir tool al-yawm bisabab al-shughl} &
\ar{Ana kunt muttawattirah tool al-yawm bisabab al-shughl} \\

\hline

D2: Gendered Adjectival Lexical Preference Shift &
Adjectives and degree modifiers are shifted while preserving factual content. &
\ar{Al-yawm kan mutaab shwai} &
\ar{Al-yawm kan mutaab jiddan wa mujhid} \\

\hline

D3: First-Person Self-Expression Framing &
Self-referential framing is made more expressive without changing outcomes. &
\ar{Ma kan al-mawduu muhim bil-nisbah li} &
\ar{Ma hassituh muhim bil-nisbah li shakhsiyyan} \\

\hline

D4: Politeness and Hedging Pragmatic Shift &
Assertions are softened using hedging while preserving stance. &
\ar{Hadha al-qarar ghalat} &
\ar{Yumkin hadha al-qarar yakun ghalat} \\

\hline

D5: Emotion and Affect Density Amplification &
Emotional expression is intensified while preserving sentiment direction. &
\ar{Hasit bil-daght} &
\ar{Hasit bidaght kabir wa tawtur wadih} \\

\hline
\end{tabular}
\caption{Arabic gender style-transfer datasets. Each example shows the source sentence and its corresponding transformed variant.}
\label{tab:datasets}
\end{table}

\section{Conditional Diffusion Model}

\begin{figure}
    \centering
    \includegraphics[width=0.8\linewidth]{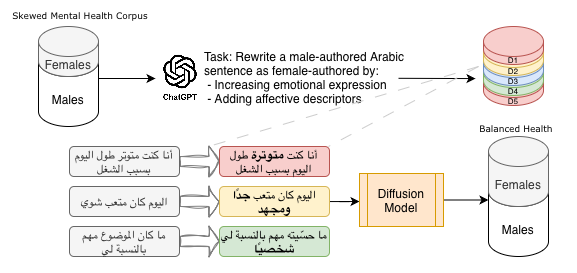}
    \caption{Pipeline for synthetic data augmentation using diffusion models.}
    \label{fig:placeholder}
\end{figure}

We adopt the text style transfer diffusion model proposed by \cite{gong2022diffuseq}, using an Arabic-compatible tokenizer \cite{antoun2020arabert}. 

\subsection{Training and Inference}
Following \cite{gong2022diffuseq}, we employ a simplified diffusion objective during training. Given a pair $(S, T_{\mathrm{RG}})$, where $S$ denotes the male source sentence augmented with style tokens and $T_{\mathrm{RG}}$ represents the corresponding female target sentence from one of the five generated datasets, we sample a timestep $t \in \{1, \ldots, T\}$ and add Gaussian noise to the target embeddings $Z^{T_{\mathrm{RG}}}_0$ to obtain $Z^{T_{\mathrm{RG}}}_t$ under a linear noise schedule. The noisy target embeddings are concatenated with the source embeddings $Z_S$ and passed through a diffusion transformer, which predicts a denoised target representation $Z^{\prime T_{\mathrm{RG}}}_0$. Training minimizes an $\ell_2$ loss between the original target embeddings $Z^{T_{\mathrm{RG}}}_0$ and their reconstruction $Z^{\prime T_{\mathrm{RG}}}_0$.

During inference, we initialize the target representation with Gaussian noise and iteratively denoise it while conditioning on the source embeddings. The final denoised embeddings are mapped to the nearest token embeddings using cosine similarity and decoded to form the output sequence.

\section{Experiments}
We train five diffusion models, each corresponding to one of the five generated datasets. The objective of training is to generate female-authored posts in order to mitigate the skewed gender distribution present in the original data. At the same time, it is important that the generated posts follow a distribution that closely matches that of the existing dataset. To this end, generation is conditioned on male-authored posts from the dataset. The input to the diffusion model therefore consists of male posts, and the output is the same content rewritten in a female voice, with varying degrees of lexical and semantic variation depending on the dataset.

For each of the five datasets, we adopt a 90/5/5 train, test, and validation split. We prioritize the size of the training split due to the limited size of the generated corpus. Each diffusion model is trained independently on a single dataset, implementing one style transfer configuration at a time. The source data remains identical across all models, consisting of the original male-authored posts from the CARMA dataset, while the target data corresponds to GPT-generated rewrites in the female voice according to the specific style transfer strategy of each dataset.

\subsection{Gender Annotator}
We constructed an LLM based gender annotator to analyze gender of the data. When we prompted the model to classify the output of the diffusion model as either male or female, we obtained promising results indicating the models successfully capture gendered linguistic patterns.

\section{Results and Discussion}
Table~\ref{tab:best-checkpoints-abs-path} presents the evaluation results for the five trained diffusion models across the constructed gender style transfer subtasks.

Across all runs, the selected models exhibit consistently high semantic similarity, with BERTScore F1 values ranging from approximately 0.93 to 0.95. This indicates that the diffusion-based style transfer models reliably preserve the underlying meaning of the source text across different training configurations. In contrast, surface-level overlap metrics are substantially lower. ROUGE scores remain below 0.07 across all runs, and ROUGE-2 is often close to zero, reflecting intentional lexical and structural divergence introduced by the style transfer process. This behavior aligns with the objective of rewriting male-authored content in a female voice while preserving semantic content.

BLEU scores vary considerably across runs, ranging from single-digit values to nearly 57. This variability suggests differing degrees of conservativeness in the generated outputs: higher BLEU scores correspond to milder transformations that remain closer to the reference text, whereas lower BLEU scores indicate stronger stylistic rewriting. Together, these results highlight a clear distinction between semantic fidelity and lexical similarity, consistent with the goals of style transfer rather than direct paraphrasing.

Overall, the high BERTScore values indicate that the generated texts largely retain their original semantic content, reducing the risk of topic drift or meaning distortion during gender transformation. At the same time, the relatively low ROUGE scores and variable BLEU values suggest that the models introduce meaningful surface-level changes, such as gendered morphology, pronoun usage, and stylistic cues, rather than performing trivial copying. This balance supports the use of the generated outputs as semantically faithful yet stylistically distinct synthetic data, which is essential for mitigating gender imbalance while preserving label validity in downstream mental health modeling tasks.

\begin{table}
\centering
\begin{tabular}{lccccc}
\hline
\textbf{Model} & \textbf{R-1} & \textbf{R-2} & \textbf{R-L} & \textbf{BERTScore} & \textbf{BLEU} \\
\hline
05 & 0.047 & 0.000 & 0.047 & 0.930 & 11.84 \\
03 & 0.053 & 0.009 & 0.053 & \textbf{0.950} & 22.65 \\
04 & 0.022 & 0.000 & 0.022 & 0.925 & 9.00 \\
01 & 0.060 & 0.023 & 0.060 & 0.949 & \textbf{56.99} \\
02 & 0.068 & 0.015 & 0.068 & 0.939 & 18.17 \\
\hline
\end{tabular}
\caption{Best-performing checkpoint from each DiffuSeq run, selected by maximum BERTScore F1. R-1, R-2, and R-L denote ROUGE-1, ROUGE-2, and ROUGE-L respectively.}
\label{tab:best-checkpoints-abs-path}
\end{table}

\section{Conclusion}
In this study, we investigate the generation of text-based synthetic data as a means of addressing gender imbalance in mental health analysis. Using the CARMA Arabic corpus, which exhibits a substantial skew toward male-authored content, we propose a diffusion-based, pretraining-free approach to generate female-authored samples by conditioning directly on existing male posts. By framing synthetic data generation as a style transfer problem, our method preserves the semantic content of the original text while introducing controlled stylistic transformations that reflect female linguistic expression in Arabic.

Through the construction of five datasets capturing varying levels of linguistic and semantic gender expression, and the training of corresponding diffusion models, we demonstrate that high-quality synthetic data can be produced without relying on pretrained large language models. Quantitative evaluations show consistently high semantic similarity between source and generated text, alongside meaningful surface-level divergence, indicating that the generated samples are both faithful to the original content and stylistically distinct. These properties are particularly important for mental health applications, where preserving label validity while mitigating demographic bias is critical. Overall, our results suggest that diffusion-based style transfer provides a viable and flexible alternative for synthetic data generation aimed at reducing gender bias in low-resource and sensitive domains.

\section*{Limitations}
This study has several limitations. First, we focus on Arabic content due to the availability of domain and linguistic expertise within the research team. However, the proposed methodology is not language-specific, and we encourage future work to extend it to English and other languages. Framing synthetic data generation for mental health as a style transfer problem is sufficiently flexible to support such extensions. In addition, while this work concentrates on gender-based style transfer, the same approach can be readily applied to other forms of bias, such as dialectal variation or social status.

\appendix
\section{Appendix: Supplementary Material}
\label{sec:appendix}

Additional implementation details, hyperparameter configurations, and qualitative examples of generated samples can be found in the supplementary materials.

\bibliographystyle{apalike}
\bibliography{references}

\end{document}